%% file: main.tex
\documentclass[letterpaper, 10pt, conference]{ieeeconf}
\IEEEoverridecommandlockouts

\usepackage[utf8]{inputenc}
\usepackage[T1]{fontenc}
\usepackage{amsmath,amssymb,amsfonts}
\usepackage{graphicx}
\usepackage{booktabs}
\usepackage{multirow}
\usepackage{tabularx}
\usepackage{array}
\usepackage{xcolor}
\usepackage{colortbl}
\newcolumntype{Y}{>{\centering\arraybackslash}X}
\definecolor{ecorow}{HTML}{E4EDF8}
\definecolor{hugsimhead}{HTML}{EFEFEF}
\usepackage{pgfplots}
\pgfplotsset{compat=1.17}
\usepackage{tikz}
\usepackage{algorithm}
\usepackage{algpseudocode}
\usepackage{subcaption}
\usepackage{cuted}
\usepackage{capt-of}
\usepackage{cite}
\usepackage{url}
\usepackage{xurl}
\usepackage[hidelinks]{hyperref}
\hypersetup{pdfauthor={Brayden Zhang, Mahsa Golchoubian, Igor Gilitschenski, Boris Ivanovic, Kashyap Chitta}, pdftitle={Guiding End-to-End Driving Models with Endpoint-Constrained Trajectory Optimization}}

\makeatletter
\def\tabcmp@qq{??}
\newcommand{\bcmp}[2]{\edef\tabcmp@a{#1}\edef\tabcmp@b{#2}%
  \ifx\tabcmp@a\tabcmp@qq #1\else\ifx\tabcmp@b\tabcmp@qq #1\else
  \ifdim #1pt>#2pt \textbf{#1}\else #1\fi\fi\fi}
\def\tabcmp@sdpm#1{\edef\tabcmp@a{#1}\ifx\tabcmp@a\tabcmp@qq\else$\pm$#1\fi}
\newcommand{\sdpm}[1]{\tabcmp@sdpm{#1}}
\makeatother

\newcommand{\REVWHY}[1]{}
\usepackage[normalem]{ulem}

\input{results/numbers}

\newcommand{\bx}{\mathbf{x}}\newcommand{\bp}{\mathbf{p}}
\newcommand{\eco}{ECO}
\newcommand{\hd}{HD-Score}

\newcommand{\boldparagraph}[1]{\vspace{2pt}\noindent\textbf{#1.}}

\title{\LARGE \bf
Guiding End-to-End Driving Models with\\ Endpoint-Constrained Trajectory Optimization}

\author{Brayden Zhang$^{1}$, Mahsa Golchoubian$^{1,2}$, Igor Gilitschenski$^{1,2}$, Boris Ivanovic$^{3}$, and Kashyap Chitta$^{4,5*}$%
\thanks{$^{1}$University of Toronto, $^{2}$Vector Institute, $^{3}$NVIDIA Research,$^{4}$ELLIS Institute T\"ubingen, $^{5}$KE:SAI. $^{*}$Work done in part while at NVIDIA Research. Contact: \href{mailto:brayden.zhang@mail.utoronto.ca}{brayden.zhang@mail.utoronto.ca}}%
\thanks{Project page: \href{https://brayden-zhang.github.io/guiding-e2e-eco/}{brayden-zhang.github.io/guiding-e2e-eco}}}

\makeatletter\def\@IEEEaftertitletext{\vspace{-4\baselineskip}}\makeatother

\begin{document}
\maketitle
\input{figures/fig_teaser}
\thispagestyle{empty}
\pagestyle{empty}

\input{sections/00_abstract}
\input{sections/01_introduction}
\input{sections/02_related_work}
\input{sections/03_analysis}
\input{sections/04_method}
\input{sections/05_experiments}

\input{sections/05b_posttraining}
\input{sections/06_conclusion}
\input{sections/07_acknowledgments}

\bibliographystyle{IEEEtran}
\bibliography{references}

\end{document}

%% file: results/numbers.tex
\newcommand{\dDrivorfixEcohPanelVsDrivorfixRawPanel}{+1.6}
\newcommand{\dEcohPanelEasy}{+36.2}
\newcommand{\dEcohPanelExtreme}{+2.7}
\newcommand{\dEcohPanelHard}{+1.2}
\newcommand{\dEcohPanelMedium}{+16.1}
\newcommand{\dEcohPanelVsRawPanel}{+12.9}
\newcommand{\dEcohPanelVsRawPanelRel}{71}
\newcommand{\dLtfEcohPanelVsLtfRawPanel}{+2.8}
\newcommand{\dOlclEcohEditAde}{-0.03}
\newcommand{\dOlclEcohEditFde}{+0.00}
\newcommand{\dOlclUniadEcohEditAde}{+0.03}
\newcommand{\dOlclUniadEcohEditFde}{+0.00}
\newcommand{\dOlclUniadNuscDiff}{+9.5}
\newcommand{\dOlclVavamNuscDiff}{+16.5}
\newcommand{\dUniadEcohPanelVsUniadRawPanel}{+5.6}
\newcommand{\dVadEcohPanelVsVadRawPanel}{+1.7}
\newcommand{\nBurdenCostBase}{101.2}
\newcommand{\nBurdenCostEco}{92.6}
\newcommand{\nBurdenCostRel}{8.5}
\newcommand{\nBurdenFracLower}{83}
\newcommand{\nBurdenPairs}{1,500}
\newcommand{\nBurdenSatBase}{20.1}
\newcommand{\nBurdenSatEco}{12.9}
\newcommand{\nDrivorFixEcoHPanelCOM}{96}
\newcommand{\nDrivorFixEcoHPanelHD}{36.2}
\newcommand{\nDrivorFixEcoHPanelHDsd}{0.3}
\newcommand{\nDrivorFixEcoHPanelKitti}{21.6}
\newcommand{\nDrivorFixEcoHPanelNC}{62}
\newcommand{\nDrivorFixEcoHPanelNusc}{48.0}
\newcommand{\nDrivorFixEcoHPanelPanda}{38.7}
\newcommand{\nDrivorFixEcoHPanelRC}{46.0}
\newcommand{\nDrivorFixEcoHPanelTTC}{58}
\newcommand{\nDrivorFixEcoHPanelWaymo}{43.9}
\newcommand{\nDrivorFixEcoHTrainHD}{38.8}
\newcommand{\nDrivorFixRawPanelCOM}{97}
\newcommand{\nDrivorFixRawPanelHD}{34.6}
\newcommand{\nDrivorFixRawPanelHDsd}{0.8}
\newcommand{\nDrivorFixRawPanelKitti}{18.1}
\newcommand{\nDrivorFixRawPanelNC}{63}
\newcommand{\nDrivorFixRawPanelNusc}{47.2}
\newcommand{\nDrivorFixRawPanelPanda}{38.9}
\newcommand{\nDrivorFixRawPanelRC}{44.0}
\newcommand{\nDrivorFixRawPanelTTC}{58}
\newcommand{\nDrivorFixRawPanelWaymo}{39.0}
\newcommand{\nEcoHPanelCOM}{97}
\newcommand{\nEcoHPanelFracImproved}{71}
\newcommand{\nEcoHPanelGainMedian}{+3.3}
\newcommand{\nEcoHPanelHD}{31.0}
\newcommand{\nEcoHPanelHDsd}{0.6}
\newcommand{\nEcoHPanelKitti}{22.2}
\newcommand{\nEcoHPanelNC}{54}
\newcommand{\nEcoHPanelNusc}{39.6}
\newcommand{\nEcoHPanelPanda}{30.3}
\newcommand{\nEcoHPanelRC}{45.2}
\newcommand{\nEcoHPanelTTC}{46}
\newcommand{\nEcoHPanelWaymo}{40.0}
\newcommand{\nEcoHTrainHD}{32.1}
\newcommand{\nLawArms}{10}
\newcommand{\nLawP}{0.0479}
\newcommand{\nLawRho}{0.64}
\newcommand{\nLtfEcoHPanelCOM}{100}
\newcommand{\nLtfEcoHPanelHD}{20.7}
\newcommand{\nLtfEcoHPanelHDsd}{0.3}
\newcommand{\nLtfEcoHPanelKitti}{8.2}
\newcommand{\nLtfEcoHPanelNC}{38}
\newcommand{\nLtfEcoHPanelNusc}{31.9}
\newcommand{\nLtfEcoHPanelPanda}{21.6}
\newcommand{\nLtfEcoHPanelRC}{33.2}
\newcommand{\nLtfEcoHPanelTTC}{36}
\newcommand{\nLtfEcoHPanelWaymo}{29.5}
\newcommand{\nLtfRawPanelCOM}{100}
\newcommand{\nLtfRawPanelHD}{17.9}
\newcommand{\nLtfRawPanelHDsd}{1.1}
\newcommand{\nLtfRawPanelKitti}{3.7}
\newcommand{\nLtfRawPanelNC}{44}
\newcommand{\nLtfRawPanelNusc}{17.8}
\newcommand{\nLtfRawPanelPanda}{22.6}
\newcommand{\nLtfRawPanelRC}{31.6}
\newcommand{\nLtfRawPanelTTC}{39}
\newcommand{\nLtfRawPanelWaymo}{34.5}
\newcommand{\nOlclEcohAde}{1.66}
\newcommand{\nOlclEcohDeficit}{0}
\newcommand{\nOlclEcohFde}{2.42}
\newcommand{\nOlclEcohHD}{39.6}
\newcommand{\nOlclRawAde}{2.53}
\newcommand{\nOlclRawDeficit}{44}
\newcommand{\nOlclRawFde}{3.27}
\newcommand{\nOlclUniadAde}{1.86}
\newcommand{\nOlclUniadDeficit}{16}
\newcommand{\nOlclUniadEcohAde}{2.02}
\newcommand{\nOlclUniadEcohDeficit}{0}
\newcommand{\nOlclUniadEcohFde}{2.61}
\newcommand{\nOlclUniadEcohHD}{37.3}
\newcommand{\nOlclUniadFde}{2.25}
\newcommand{\nRawPanelCOM}{14}
\newcommand{\nRawPanelHD}{18.1}
\newcommand{\nRawPanelHDsd}{0.5}
\newcommand{\nRawPanelKitti}{4.0}
\newcommand{\nRawPanelNC}{52}
\newcommand{\nRawPanelNusc}{23.1}
\newcommand{\nRawPanelPanda}{23.1}
\newcommand{\nRawPanelRC}{40.9}
\newcommand{\nRawPanelTTC}{38}
\newcommand{\nRawPanelWaymo}{25.8}
\newcommand{\nRawTrainHD}{20.0}
\newcommand{\nUniadEcoHPanelCOM}{100}
\newcommand{\nUniadEcoHPanelHD}{32.4}
\newcommand{\nUniadEcoHPanelKitti}{8.1}
\newcommand{\nUniadEcoHPanelNC}{72}
\newcommand{\nUniadEcoHPanelNusc}{37.3}
\newcommand{\nUniadEcoHPanelPanda}{48.0}
\newcommand{\nUniadEcoHPanelRC}{39.8}
\newcommand{\nUniadEcoHPanelTTC}{63}
\newcommand{\nUniadEcoHPanelWaymo}{28.5}
\newcommand{\nUniadRawPanelCOM}{72}
\newcommand{\nUniadRawPanelHD}{26.8}
\newcommand{\nUniadRawPanelHDsd}{0.8}
\newcommand{\nUniadRawPanelKitti}{2.9}
\newcommand{\nUniadRawPanelNC}{69}
\newcommand{\nUniadRawPanelNusc}{27.8}
\newcommand{\nUniadRawPanelPanda}{42.6}
\newcommand{\nUniadRawPanelRC}{36.7}
\newcommand{\nUniadRawPanelTTC}{57}
\newcommand{\nUniadRawPanelWaymo}{27.5}
\newcommand{\nVadEcoHPanelCOM}{100}
\newcommand{\nVadEcoHPanelHD}{17.4}
\newcommand{\nVadEcoHPanelHDsd}{0.4}
\newcommand{\nVadEcoHPanelKitti}{1.9}
\newcommand{\nVadEcoHPanelNC}{52}
\newcommand{\nVadEcoHPanelNusc}{28.6}
\newcommand{\nVadEcoHPanelPanda}{24.9}
\newcommand{\nVadEcoHPanelRC}{29.4}
\newcommand{\nVadEcoHPanelTTC}{39}
\newcommand{\nVadEcoHPanelWaymo}{11.0}
\newcommand{\nVadRawPanelCOM}{100}
\newcommand{\nVadRawPanelHD}{15.7}
\newcommand{\nVadRawPanelHDsd}{0.9}
\newcommand{\nVadRawPanelKitti}{1.1}
\newcommand{\nVadRawPanelNC}{55}
\newcommand{\nVadRawPanelNusc}{23.8}
\newcommand{\nVadRawPanelPanda}{24.4}
\newcommand{\nVadRawPanelRC}{26.6}
\newcommand{\nVadRawPanelTTC}{43}
\newcommand{\nVadRawPanelWaymo}{8.2}
\newcommand{\nVamBParams}{357M}
\newcommand{\nVamLParams}{1.4B}
\newcommand{\nVamSParams}{204M}
\newcommand{\sDrivorfixEcohPanelVsDrivorfixRawPanel}{0.9}
\newcommand{\sEcohPanelVsRawPanel}{0.8}
\newcommand{\sLtfEcohPanelVsLtfRawPanel}{1.1}
\newcommand{\sUniadEcohPanelVsUniadRawPanel}{1.1}
\newcommand{\sVadEcohPanelVsVadRawPanel}{1.0}

%% file: figures/fig_teaser.tex
\begin{strip}
    \centering
    \includegraphics[width=0.996\textwidth]{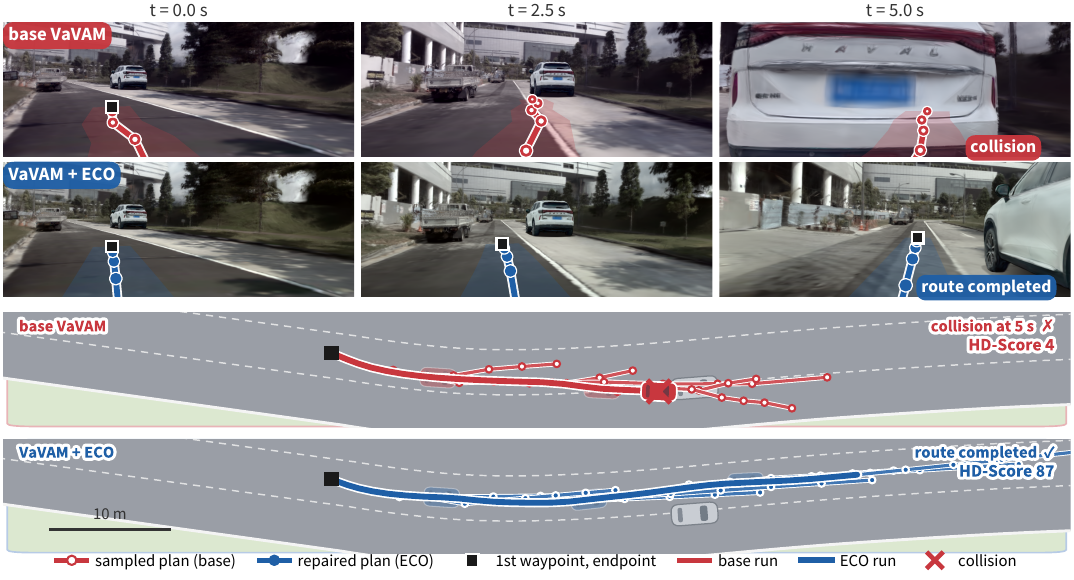}
    \captionof{figure}{\textbf{Endpoint-Constrained Optimization (\eco{})} in HUGSIM. Top: Front-camera views comparing the base VaVAM policy against VaVAM + ECO. At $t=0$ the sampled base plan (red) exhibits lateral inconsistencies. ECO retains the endpoint and starts from the executed state while optimizing the waypoints between them into a smooth trajectory (blue). In the resulting closed-loop rollouts, base VaVAM later collides, whereas VaVAM + ECO completes the route. Bottom: Bird's-eye view of the same rollouts, where ECO allows the vehicle to complete the route, achieving a higher HD-Score.}\label{fig:teaser}
\end{strip}

%% file: sections/00_abstract.tex
\begin{abstract}
End-to-end driving policies are commonly trained through open-loop behavior cloning, yet they must ultimately operate in closed-loop when deployed on a vehicle, creating a fundamental mismatch between training and execution. Beyond the commonly studied effects of covariate shift and causal confusion, we identify a complementary factor for this open-loop/closed-loop gap: waypoint-based supervision and displacement metrics do not ensure that the intermediate trajectory is physically coherent or easy for the controller to track. We observe that these inconsistencies concentrate primarily at intermediate waypoints, while the predicted endpoint remains comparatively reliable. Based on this observation, we introduce Endpoint-Constrained Optimization (ECO), a lightweight post-processing layer that anchors the trajectory to the vehicle's executed history, preserves the policy's predicted endpoint, and reshapes the intermediate waypoints to improve feasibility. ECO requires no map, privileged simulator state, or additional training, and can be inserted between a broad range of waypoint-emitting policies and their controllers. Across two closed-loop simulators, it improves the aggregate closed-loop score of all six evaluated generative and regression-based driving policies, and the gains tend to increase with how often the base plans violate motion limits. On HUGSIM, \eco{} improves VaVAM from \nRawPanelHD{} to \nEcoHPanelHD{} \hd{} (+\dEcohPanelVsRawPanelRel\%), achieving 1st place on the HUGSIM Closed-Loop Driving Challenge. Similarly, on AlpaSim, \eco{} increases the scene scores of VaVAM and DiffusionDrive by 123\% and 22\%, respectively. These results show that for a broad collection of end-to-end driving models, repairing the intermediate geometry of predicted trajectories without changing the policy's predicted endpoint can substantially improve closed-loop performance.
\end{abstract}

%% file: sections/01_introduction.tex
\section{Introduction}

\label{sec:intro}

Driving policies are increasingly trained end-to-end to predict future trajectories from sensor observations and navigation commands~\cite{Chen2024PAMI}.
This is most commonly done through open-loop behavior cloning, where predicted trajectories are compared against logged human driving.
In closed-loop evaluation, however, the vehicle executes actions derived from these predictions autoregressively, changing its future observations and subsequent predictions.
As a result, a policy may achieve strong open-loop trajectory metrics yet fail during rollouts in closed-loop environments, illustrating the well-known open-loop/closed-loop (OL/CL) gap~\cite{Codevilla2018ECCV}.

Most existing work on this gap has focused on covariate shift and causal confusion~\cite{CausalConfusionNeurIPS2019, karkus2025beyondbc} as potential causes.
However, a complementary factor is that open-loop training uses waypoint-based representations that reward agreement with logged positions without explicitly enforcing the physical plausibility of the output trajectories.
Training loss functions are typically applied to all predicted waypoints at an equal scale, leading to a larger learning signal for distant endpoints than nearer intermediate waypoints.
Open-loop metrics like average displacement error (ADE) look at waypoint positions in isolation, so they miss the accelerations, jerks, and discontinuities among the waypoints as well as between the vehicle's historical states and waypoint predictions.
Other metrics like final displacement error (FDE) ignore the entire path except the endpoint. As a result, a plan may score strongly in open-loop even if its intermediate waypoints are hard for the controller to follow and extract actions from.

We therefore introduce \textbf{Endpoint-Constrained Optimization} \textbf{(\eco{})}, a training-free post-processor that improves a policy's executability by preserving the policy's high-level plan while refining the intermediate trajectory.
Specifically, we apply a lightweight optimization on the sequence of historical and predicted waypoints, with the history and end poses kept fixed, encouraging physical plausibility.
ECO consumes and returns the same fixed-period waypoint representation, so it can be inserted between any waypoint-emitting policy and its controller, whether the policy uses a generative~\cite{vavam2025, liao2025diffusiondrive} or regression-based~\cite{hu2023uniad, jiang2023vad, Chitta2023PAMI, kirby2026drivor} head.
Across two simulators~\cite{zhou2024HUGSIM, alpasim2025} and five datasets (nuScenes~\cite{nuscenes}, Waymo~\cite{WaymoOpenDataset}, KITTI-360~\cite{Liao2022PAMI}, PandaSet~\cite{Xiao2021ITSC}, and the AlpaSim NuRec set~\cite{alpasim2025}), \eco{} increases the aggregate closed-loop score of every evaluated policy. These improvements are significant enough to place first in the HUGSIM Closed-Loop Driving Challenge~\cite{realadsim2025challenge} and require no simulator-specific hyperparameter re-tuning, despite differences in vehicle dynamics, controllers, and evaluation settings.

Our contributions are:
\begin{itemize}

\item An analysis of the OL/CL gap showing that standard displacement metrics can miss physically inconsistent waypoint sequences, with most violations occurring in the middle of the trajectory rather than at its endpoints.

\item \eco{}, a training-free trajectory optimization layer that keeps the executed history and the policy's endpoint fixed and repairs the trajectory between them.

\item A set of controlled closed-loop studies that isolate why this refinement works and when it transfers.%

\end{itemize}

%% file: sections/02_related_work.tex
\section{Related Work}
\label{sec:related}
\input{figures/fig_pipeline}

\boldparagraph{Closed-loop evaluation of end-to-end driving}
Photorealistic neural simulators such as HUGSIM~\cite{zhou2024HUGSIM} and AlpaSim~\cite{alpasim2025} now make it significantly easier to study how policies behave under online interaction, rather than only through offline trajectory metrics.
Even on these new photorealistic simulators, our work shows that behavior cloning for end-to-end driving suffers from a discrepancy between offline evaluation and actual closed-loop performance, in line with earlier work~\cite{Codevilla2018ECCV, ross2011dagger, CausalConfusionNeurIPS2019}.
In particular, recent end-to-end driving policies increasingly use generative planning heads based on diffusion~\cite{liao2025diffusiondrive} and flow matching~\cite{lipman2023flow} to model complex multi-modal behaviors.
We observe a large open-loop/closed-loop gap for this now predominant generative planning paradigm.

\boldparagraph{Improving closed-loop robustness}
Existing approaches address closed-loop failures at several levels.
Interactive data collection~\cite{ross2011dagger} (either in real life or simulation) and simulation-based closed-loop fine-tuning~\cite{wagenmaker2025steering} modify the policy using feedback from expert labels or closed-loop rewards, respectively.
Other work analyzes biases in open-loop training and evaluation to improve closed-loop robustness.
For example, driving models can learn shortcuts such as target point following~\cite{Jaeger2023ICCV}, which can be mitigated in open-loop training to improve closed-loop performance.
In a similar spirit, we analyze the impact of physical implausibility stemming from open-loop training using waypoint prediction losses.
Our proposed \eco{} module leaves the policy's training unchanged and instead operates at the policy--controller interface, improving closed-loop executability.

\boldparagraph{Trajectory refinement and post-processing}
Trajectory optimization has long been used to refine candidate motions under geometric and dynamical constraints.
Classical methods include CHOMP~\cite{ratliff2009chomp} and TrajOpt~\cite{schulman2014trajopt}, while driving-specific approaches include Frenet-frame generation~\cite{werling2010frenet} and path--velocity decomposition~\cite{kant1986pathvelocity}.
More recent driving systems also account explicitly for downstream execution.
For example, PDM-Closed~\cite{Dauner2023CORL} simulates rule-based trajectory proposals through a vehicle model and tracking controller before selecting among them, while safety filters and control barrier functions intervene at the control level to enforce constraints~\cite{ames2019cbf,hsu2024safetyfilter}.
Other approaches modify or select learned trajectories, including safety-oriented planners and proposal-selection schemes~\cite{vitelli2022safetynet, li2024hydramdp}, or apply conditioning/constraints to guide diffusion-based trajectory models~\cite{kondo2024cgd, xiao2025safediffuser, janner2022diffuser}.
Closer to our setting, Hydra-NeXt~\cite{li2025hydranext} learns a trajectory-refinement module jointly with trajectory and control prediction to improve kinematic consistency, while TOAD~\cite{xu2026toad} performs test-time trajectory search using a learned trajectory scorer.
In contrast, \eco{} requires neither additional training nor a learned scorer, and operates only on the policy's emitted trajectory and execution history while preserving the predicted endpoint. Despite this simplicity, it shows strong effectiveness across simulators and policies.

%% file: figures/fig_pipeline.tex
\begin{figure*}[t]
    \centering
    \includegraphics[width=0.95\textwidth]{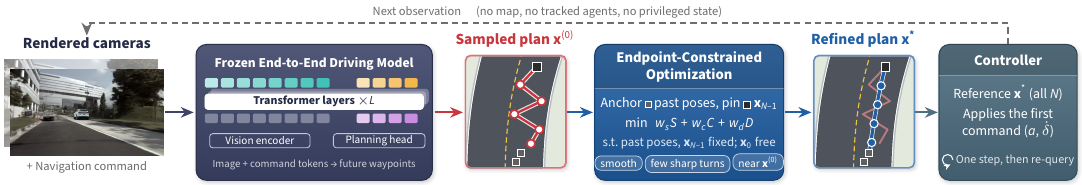}
    \caption{\textbf{Pipeline.} The frozen end-to-end driving model receives camera inputs from the simulator and samples a trajectory for the next $N$ steps. ECO anchors $K=2$ past poses and the last predicted waypoint, then solves a trajectory optimization problem to refine the waypoints between them (Sec.~\ref{sec:method}). The controller then tracks the refined plan.}
    \label{fig:pipeline}
\end{figure*}

%% file: sections/03_analysis.tex
\section{Problem Formulation}
\label{sec:problem}

\boldparagraph{Closed-loop policy--controller interface}
A pre-trained end-to-end driving policy receives images and a navigation command from its environment and predicts $N$ waypoints at $\Delta t$  intervals in the ego frame. A trajectory-tracking controller~\cite{li2004ilqr} uses the planned trajectory as its reference to extract a control sequence, applies the first action command in this sequence to reach a new environment state, and then queries the policy again.

\boldparagraph{Predicted waypoint sequence}
The policy emits a predicted plan $\bx^{(0)}=(\bx^{(0)}_0,\ldots,\bx^{(0)}_{N-1})$ of $N$ waypoints at some interval $\Delta t$, where $N$ is determined by the underlying policy.
Each waypoint is $\bx_i=(\bp_i,\theta_i)$, with position $\bp_i=(x_i,y_i)$ and heading $\theta_i$.
For policies that predict positions only, we define $\theta_i$ as the heading of the segment from $\bp_{i-1}$ to $\bp_i$, with $\bp_{-1}$ at the ego origin. We write the vehicle's executed poses in the same $(\bp,\theta)$ form as $\mathbf{h}_{-K},\ldots,\mathbf{h}_{-1}$, with the current pose $\mathbf{h}_0=(0,0,0)$ at the origin of the ego frame.

\boldparagraph{Post-inference trajectory optimization}
We apply an optimization step after inference and before control that maps the predicted plan to an optimized plan $\bx^\star$, which becomes the controller's reference with the same horizon and timing.
The base policy remains unchanged, and the optimization itself only sees the emitted plan and executed ego history.

\boldparagraph{Problem}
Which parts of the emitted plan should such a post-processor be allowed to change? Sec.~\ref{sec:analysis} shows that motion-limit violations concentrate in the plan interior; we therefore ask whether repairing the intermediate waypoints $\bx_{0:N-2}$, while keeping the last waypoint fixed, improves closed-loop performance in terms of route completion, driving safety, and comfort. Sec.~\ref{sec:method} instantiates this as \eco{}.

%% file: sections/04_method.tex
\section{Endpoint-Constrained Optimization}
\label{sec:method}
\label{sec:smoothing}

\eco{} is placed between the policy and controller (Fig.~\ref{fig:pipeline}). Before optimization, we prepend the two most recently executed poses $\mathbf{h}_{-2:-1}$ and the current pose $\mathbf{h}_0=(0,0,0)$ to the predicted trajectory. We hold these historical poses and the final predicted waypoint $\bx_{N-1}$ fixed, while optimizing the remaining waypoints $\bx_{0:N-2}$. Each historical pose $\mathbf{h}_{-k}$ represents the vehicle pose $k\Delta t$ before the current query, with $\Delta t$ matching the prediction interval. We obtain these poses by interpolating the executed ego trajectory and transforming them into the current ego frame. At the beginning of an episode, when sufficient history is not yet available, we anchor only the executed poses that exist, starting with the current pose.
Because the objective is evaluated over the combined executed-and-predicted trajectory, its finite differences encourage a smooth transition into the new plan.

\boldparagraph{Objective}
We optimize the intermediate waypoints using an objective that balances trajectory smoothness, turn sharpness, and fidelity to the predicted plan. These terms are evaluated on the extended trajectory $\mathbf{z}=(\mathbf{h}_{-K},\ldots,\mathbf{h}_0,\bx_0,\ldots,\bx_{N-1})$ (Table~\ref{tab:cost_terms}), in which the anchored poses $\mathbf{h}_{-K:0}$ are fixed parameters, and the deviation term is applied on the plan alone. We optimize:
\begin{align}
\bx^\star = \arg\min_{\bx}\;&
w_s S(\mathbf{z}) + w_c C(\mathbf{z}) + w_d D(\bx,\bx^{(0)}) \label{eq:cost}\\
\text{s.t.}\;&
\bx_{N-1}=\bx^{(0)}_{N-1}. \label{eq:endpoint}
\end{align}

\input{figures/tab_cost_terms}
Table~\ref{tab:cost_terms} defines the three objective terms, with $\mathbf{z}_i=\mathbf{h}_{i+1}$ for $i<0$ and $\mathbf{z}_i=\bx_i$ for $i\geq 0$.
The smoothness term $S$ penalizes second differences in position and heading, discouraging abrupt changes between consecutive waypoints; angular differences are wrapped to $[-\pi,\pi]$. The turn penalty $C$ penalizes sharp turns using the cross product $c_i$ of consecutive trajectory segments. We apply this penalty only when $|c_i|>c_0=0.5   $ $\mathrm{ m}^2$ and both segments are longer than $0.1$~m, avoiding unstable penalties near stationary points. Finally, the deviation term $D$ discourages the optimized trajectory from moving too far from the policy's original prediction. Positions and headings are jointly optimized.

\boldparagraph{Optimization}
We solve Eq.~\eqref{eq:cost} with L-BFGS-B~\cite{L-BFGS-B}, initialized at the predicted plan. On a Xeon Gold 6238R CPU, the algorithm takes a median of 31 iterations and 20.7 ms to solve. This is comfortably within a typical end-to-end planner's inference budget, allowing parallelized execution and pipelining (e.g., the inference of VaVAM~\cite{vavam2025} takes 44.7 ms forward pass on an L40S GPU). It also comfortably fits within the assigned 250\,ms and 100\,ms control periods for the HUGSIM and AlpaSim simulators, respectively.

%% file: figures/tab_cost_terms.tex
\begin{table}[t]
\centering\scriptsize
\setlength{\tabcolsep}{2pt}
\renewcommand{\arraystretch}{1.1}
\caption{\textbf{Components of Eq.~\eqref{eq:cost}.}
Indices run over the extended trajectory $\mathbf{z}$ (Sec.~\ref{sec:method}).}
\label{tab:cost_terms}
\begin{tabularx}{\columnwidth}{@{}>{\raggedright\arraybackslash}p{1.3cm}X@{}}
\toprule
Term & Definition \\
\midrule
Smoothness\newline $S(\mathbf{z})$ &
$\displaystyle\sum_{i=-K-1}^{N-3}\Big[\left|\bp_{i+2}-2\bp_{i+1}+\bp_i\right|^2$\newline $\displaystyle\qquad\quad+\operatorname{wrap}\!\left(\theta_{i+2}-2\theta_{i+1}+\theta_i\right)^2\Big]$ \\[1.2ex]
Turn penalty\newline $C(\mathbf{z})$ &
$\displaystyle\sum_{i=-K}^{N-2}\mathbf{1}_i\,(|c_i|-c_0)^2,\quad c_i=(\bp_i-\bp_{i-1})\times(\bp_{i+1}-\bp_i)$ \\[1.2ex]
Deviation\newline $D(\bx,\bx^{(0)})$ &
$\displaystyle\sum_{i=0}^{N-1}\left|\bx_i-\bx_i^{(0)}\right|^2$ \\[1.2ex]
\bottomrule
\end{tabularx}
\end{table}

%% file: sections/05_experiments.tex
\section{Experiments}
\label{sec:experiments}
\input{tables/tab_olcl}

Our experiments address three questions: where trajectory inconsistencies arise, whether ECO improves closed-loop driving across policies and simulators, and which components drive the gains. We first describe the experimental setup, then analyze trajectory errors, evaluate ECO, and conclude with controlled ablations.
\subsection{Experimental Setup}
\label{sec:setup}

\boldparagraph{Benchmarks and evaluation splits}
Our primary closed-loop benchmark is HUGSIM~\cite{zhou2024HUGSIM}, a photorealistic closed-loop simulator built from 3D Gaussian Splatting~\cite{kerbl2023gaussians} reconstructions of real-world driving logs from nuScenes~\cite{nuscenes}, Waymo~\cite{WaymoOpenDataset}, KITTI-360~\cite{Liao2022PAMI}, and PandaSet~\cite{Xiao2021ITSC}. HUGSIM provides reactive traffic across four difficulty levels, with the hard and extreme modes introducing aggressive actors that may perform adversarial, collision-seeking maneuvers. The HUGSIM dataset has two splits: a 345-scene ``train'' split released before the associated closed-loop driving challenge, and a 185-scene ``test'' split released afterward. While \eco{} does not involve any training, we maintain this naming convention throughout the results. For ablation studies, we use the nuScenes subset of the HUGSIM test split. We additionally evaluate in AlpaSim~\cite{alpasim2025}, a neural reconstruction-based closed-loop simulator from NVIDIA. It runs its own traffic simulation and vehicle dynamics, and tracks the policy's plan with a model-predictive controller rather than HUGSIM's iLQR-based controller. For evaluation, following~\cite{dacol2026opted}, we use the 441-scene NuRec validation set of the AlpaSim end-to-end driving challenge.

\input{figures/fig_horizon}

\boldparagraph{Policies and controllers}
Our primary analysis uses VaVAM~\cite{vavam2025}, which combines a video-generative backbone pre-trained on OpenDV-2K~\cite{yang2024genad} with a flow-matching~\cite{lipman2023flow} action head post-trained on nuPlan~\cite{nuplan} and nuScenes~\cite{nuscenes}; ``VaVAM'' denotes the released VaVAM-B checkpoint throughout, and VaVAM-S/-L (Fig.~\ref{fig:deficitlaw}) are the released smaller and larger checkpoints of the same architecture and training recipe. We evaluate \eco{} in closed loop using HUGSIM and AlpaSim with their default controllers, unless otherwise mentioned. Both the policy sampler and the traffic simulation can be stochastic: the action head samples noise at each control step, while the hard and extreme HUGSIM scenes include stochastic agent behavior.

\boldparagraph{Baselines}
We evaluate a set of diverse planning architectures with ECO. In addition to the flow-matching-based VaVAM, we also use the truncated diffusion-based DiffusionDrive~\cite{liao2025diffusiondrive} as well as several deterministic or proposal-based planners, namely DrivoR~\cite{kirby2026drivor}, UniAD~\cite{hu2023uniad}, VAD~\cite{jiang2023vad}, and Latent TransFuser (LTF)~\cite{Chitta2023PAMI}. Unless otherwise specified, all methods use the same scenes, controller, and evaluation protocol. As post-processing baselines, we replace \eco{} with alternative trajectory-shaping operators under the same endpoint constraints (Sec.~\ref{sec:ablation}).

\boldparagraph{Metrics and evaluation protocol}
Within HUGSIM, the primary \hd{} metric combines route completion $R_c$ with a per-step score in which no-collision (NC) and drivable-area (DAC) act as gates and time-to-collision (TTC) and comfort (COM) contribute to the remaining score. For the exact equations, please refer to~\cite{zhou2024HUGSIM}. For AlpaSim, following the setting of~\cite{dacol2026opted}, we use the standard metrics of the \texttt{e2e\_challenge} branch. These metrics are frozen at the first at-fault collision, off-road event, or exit from the 4 m corridor around the logged route. The scene score rewards route progress above 80\% for scenes without these failures, while m/incident is the distance driven per at-fault collision or off-road event, in meters. We also report route progress and the corresponding failure rates. For the detailed analysis of trajectory feasibility, we define a waypoint as \emph{violating a limit} when the finite differences of the predicted trajectory exceed one of the HUGSIM motion limits~\cite{zhou2024HUGSIM}: yaw rate, yaw acceleration, longitudinal acceleration, and jerk, with thresholds of 0.95 rad/s, 1.93 rad/s$^2$, $[-4.05,4.89]$ m/s$^2$, and 8.37 m/s$^3$, respectively. We define the \emph{kinematic deficit} of a plan or policy as the fraction of its waypoints that violate at least one of these limits. Note that we use these limits only for evaluation, not in Eq.~\eqref{eq:cost}, and hold them fixed across policies and simulators.

\boldparagraph{Implementation details}
We select \(w_s=3\), \(w_c=8\), and \(w_d=0.05\) using a coarse hyperparameter sweep on a small held-out subset of the HUGSIM train split. We chose the setting that reduced kinematic-limit violations while maintaining fidelity to the original predicted trajectory, and subsequently froze these weights for all reported policies, HUGSIM test experiments, and AlpaSim evaluations. These parameters are relatively insensitive to moderate value changes; scaling the smoothness, turn penalty, or deviation weight by $3\times$ or $1/3\times$ changes \hd{} by no more than 2.4 points. The only exception is when the deviation penalty is increased beyond $w_d = 0.5$ to hold the plan to the original trajectory and prevent the optimization almost entirely.

\subsection{Where Errors Arise Along Planned Trajectories}
\label{sec:analysis}
\label{sec:horizon}

\boldparagraph{Open-loop displacement is misaligned with closed-loop success}
For each predicted plan, we first calculate the mean waypoint displacement from the relevant human trajectory.
Across all the episodes driven by VaVAM in the nuScenes subset of HUGSIM, plans from collision episodes are \emph{closer} to the human driving path than those from completed episodes, with median displacements of 1.2 and 2.0\,m, respectively.
Hence, merely having good geometric agreement with the demonstration does not show whether a predicted trajectory will be easy to carry out in a closed loop.

Table~\ref{tab:olcl} shows similar patterns across two policies on the same HUGSIM nuScenes split. \eco{} changes the predicted trajectories relative to the human route by only centimeters and, for UniAD, slightly \emph{away} from it (higher ADE), while the closed-loop score on the same episodes increases by \dOlclVavamNuscDiff{} points for VaVAM and \dOlclUniadNuscDiff{} points for UniAD.

\input{tables/tab_headline}
\boldparagraph{Failure concentrates in the interior}
We next examine how trajectory errors vary along the prediction horizon. At each predicted waypoint, we place the ego vehicle's spatial footprint at the predicted pose and test for overlap with agent boxes from the corresponding future frame of the same rollout, using the final frame for waypoints beyond the episode. The intersection rate rises from 1.2\% at the first predicted waypoint to 8.2\% at 2.5\,s, before dropping to 3.5\% at the endpoint (Fig.~\ref{fig:horizon}a). We observe the same pattern for motion-limit violations, which occur predominantly at intermediate waypoints
(Fig.~\ref{fig:horizon}b), rather than at the boundaries of the prediction horizon.
Together, these analyses indicate that the largest concentration of undesirable behavior lies in the intermediate trajectory waypoints.

\subsection{\eco{} Improves Closed-Loop Driving}
\input{tables/tab_comparison}
\input{figures/fig_results_overview}
\label{sec:main}

\boldparagraph{Main results}
\eco{} substantially improves closed-loop performance on the full HUGSIM test set (Table~\ref{tab:headline}, Fig.~\ref{fig:results}). For VaVAM, the \hd{} increases from \nRawPanelHD{} to \nEcoHPanelHD, the largest gain among all tested policies. The gain is not limited to comfort:
Route completion for VaVAM increases from \nRawPanelRC{} to \nEcoHPanelRC{}. The improvement is not concentrated in a small number of scenes: \nEcoHPanelFracImproved\% of the 185 scenes improve (Fig.~\ref{fig:results}a), with a median per-scene gain of \nEcoHPanelGainMedian{}.
The magnitude varies with scene difficulty (Fig.~\ref{fig:results}b): \eco{} improves \hd{} by \dEcohPanelEasy{}, \dEcohPanelMedium{}, \dEcohPanelHard{}, and \dEcohPanelExtreme{} on easy, medium, hard, and extreme scenes, respectively. The smaller gains on the hardest scenes suggest that ECO alone cannot address all failure modes, particularly those that would require a completely different planned endpoint to solve the scenario.
The gains broadly match how much of the plan the objective can repair: gains are large for VaVAM, moderate for UniAD, and smaller for LTF, VAD, and DrivoR, whose plans already satisfy most motion limits.
Additionally, the gain is not limited to the test split; compared to existing baselines on the 345-scene train split, \eco{} improves VaVAM from \nRawTrainHD{} to \nEcoHTrainHD{} and DrivoR (w/ SimScale~\cite{tian2026simscale}) from 35.7 to \nDrivorFixEcoHTrainHD{} (Table~\ref{tab:comparison}).

\boldparagraph{Further evaluation in AlpaSim}
We next run VaVAM + \eco{} in AlpaSim~\cite{alpasim2025} (Table~\ref{tab:headline}, bottom), with both the default nonlinear MPC controller and an alternative linear MPC controller. We find similar gains as in HUGSIM, despite different controller types. Furthermore, AlpaSim contains no comfort score, yet \eco{} also improves its aggregate closed-loop metrics, providing additional evidence that the gain from \eco{} is not limited to comfort. We also test DiffusionDrive~\cite{liao2025diffusiondrive} as a second generative policy which improves along every reported AlpaSim metric with \eco{}.

\boldparagraph{Model sizes and generation steps}
\input{figures/fig_deficit_law}
Across \nLawArms{} policy and sampler variants analyzed in Fig.~\ref{fig:deficitlaw}, larger deficits correspond to larger closed-loop gains from \eco{} (Spearman $\rho=\nLawRho$, $p=\nLawP$). Policies whose waypoint sequences contain more kinematic inconsistencies leave more room for post-inference refinement.
For generative heads, ECO also provides a favorable speed--performance trade-off. A 2-step VaVAM sampler with ECO reaches 38.7 HD-Score on the nuScenes split of the HUGSIM test set, compared with 21.3 for the 20-step base sampler, while reducing per-plan latency from 65 ms to 49 ms. This suggests that trajectory repair can compensate for coarser generative sampling, enabling cheaper inference while improving closed-loop performance.

\input{tables/tab_endpoint_factorial}

\input{figures/fig_ablations}
\subsection{What Causes the Improvement?}
\label{sec:ablation}
\boldparagraph{Boundary conditions are important}
\eco{} achieves a 39.6 \hd{} when both the executed history and predicted endpoint are fixed (see Table~\ref{tab:endpoint}), compared to 23.1 for the base policy. If we instead fix the first predicted waypoint $\bx_0$ and the endpoint, but do not anchor the executed history, the score drops to 36.9. If the endpoint is not fixed, the score falls to 34.4 with history anchoring and 32.1 when $\bx_0$ is fixed. The first-waypoint-fixed variant therefore provides a substantial boost already, but the full \eco{} setup accounts for the final 2.7 points of performance.

\boldparagraph{Choice of objective}
Next, we test whether the improvement depends on the specific \eco{} objective (Fig.~\ref{fig:operator_ladder}). Keeping the same boundary conditions (executed history anchored, endpoint pinned, first predicted waypoint free), we replace \eco{} with a least-squares cubic smoothing spline (smoothing factor $s=N$), Savitzky--Golay filtering~\cite{savitzky1964smoothing} (window 5, order 2), a 3-point moving average, a constrained bicycle-model optimizer~\cite{kong2015kinematic}, and temporal ensembling of consecutive plans as in action chunking~\cite{zhao2023act}. These methods reach HD-Scores of 28.9, 31.9, 31.7, 34.8, and 27.8, respectively, compared with 39.6 for \eco{}. Simply replacing the interior before the policy-selected endpoints with any intuitive approach already recovers part of the lost closed-loop performance, although \eco{} gives the largest improvement.

\boldparagraph{Simplified tracking}
We also replay HUGSIM's controller on \nBurdenPairs{} paired base and \eco{} trajectories from the same ego state in HUGSIM. The controller cost, i.e.\ the iLQR objective (the sum over the tracking horizon of the quadratic state error to the reference and the quadratic control effort), falls by \nBurdenCostRel\%, from \nBurdenCostBase{} to \nBurdenCostEco{}, and is lower for \nBurdenFracLower\% of the pairs. The percentage of steps that reach the steering-rate limit drops from \nBurdenSatBase\% to \nBurdenSatEco\%. At the same time, the error in reaching the first predicted waypoint barely changes (base: 0.45\,m, ECO: 0.44\,m). This suggests that ECO's benefit comes from a more trackable reference over the whole horizon, and not specifically from easier tracking of the first waypoint.

\input{figures/fig_operator_gallery}

\subsection{Limitations}
\label{sec:limitations}
Because ECO fixes the policy's endpoint, it is limited in what it can correct. For example, if the endpoint is in a poor position, ECO cannot make the vehicle add a stop the policy did not predict. This matters especially for collisions: ECO does not reason about other agents (Sec.~\ref{sec:method}, Fig.~\ref{fig:pipeline}), so a separate safety layer may still be necessary.

%% file: tables/tab_olcl.tex
\begin{table}[t]
    \centering\scriptsize\setlength{\tabcolsep}{3.5pt}\renewcommand{\arraystretch}{1.1}
    \caption{\textbf{Small open-loop changes can produce large closed-loop gains.}
Results on the 34-scene nuScenes subset of the HUGSIM test set. ADE/FDE compare the plan to the ground truth, whereas paired $\Delta$ measures the change from each raw plan to its \eco{}-processed version. Deficit is the fraction of waypoints violating a motion limit. \eco{} changes paired ADE/FDE by only centimeters while improving \hd{} by \dOlclVavamNuscDiff{} for VaVAM and \dOlclUniadNuscDiff{} for UniAD. The ADE/FDE columns are computed over the plans encountered in each closed-loop rollout and therefore reflect different visited-state distributions. Paired $\Delta$ADE/$\Delta$FDE instead compares each raw plan with its ECO-processed version at the same state, isolating the direct effect of post-processing.}

    \label{tab:olcl}
    \begin{tabular}{l cc c c c}
        \toprule
        & \multicolumn{2}{c}{open loop (m)} & paired $\Delta$ after \eco{} & & closed loop \\
        \cmidrule(lr){2-3}\cmidrule(lr){4-4}\cmidrule(lr){6-6}
        Sent plan & ADE$\downarrow$ & FDE$\downarrow$ & $\Delta$ADE / $\Delta$FDE (m) & deficit\,[\%]$\downarrow$ & \hd{}$\uparrow$ \\
        \midrule
        VaVAM & \nOlclRawAde & \nOlclRawFde & -- & \nOlclRawDeficit & \nRawPanelNusc \\
        \rowcolor{ecorow} VaVAM + \eco{} & \nOlclEcohAde & \nOlclEcohFde & $\dOlclEcohEditAde{}$ / $\dOlclEcohEditFde$ & \nOlclEcohDeficit & \textbf{\nOlclEcohHD} \\
        UniAD & \nOlclUniadAde & \nOlclUniadFde & -- & \nOlclUniadDeficit & \nUniadRawPanelNusc \\
        \rowcolor{ecorow} UniAD + \eco{} & \nOlclUniadEcohAde & \nOlclUniadEcohFde & $\dOlclUniadEcohEditAde{}$ / $\dOlclUniadEcohEditFde$ & \nOlclUniadEcohDeficit & \textbf{\nOlclUniadEcohHD} \\
        \bottomrule
    \end{tabular}
\end{table}

%% file: figures/fig_horizon.tex
\begin{figure}[t]
    \centering
    \includegraphics[width=\columnwidth]{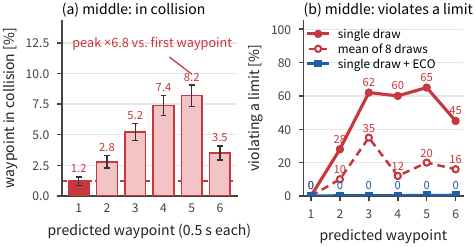}
    \caption{\textbf{Collisions and limit violations concentrate in the middle of a predicted plan.} Measured with VaVAM, on the nuScenes subset of the HUGSIM test set. (a) Fraction of predicted waypoints whose ego footprint overlaps an agent box recorded in the rollout at the frame each waypoint is meant for (i.e., waypoints in collision). (b) Waypoints violating a motion limit for a single sampled plan, for the average of eight sampled plans, and for one plan after \eco{}.}
    \label{fig:horizon}
    \label{fig:rates}
    \label{fig:dispersion}
\end{figure}

%% file: tables/tab_headline.tex
\begin{table*}[t]
    \centering\scriptsize\setlength{\tabcolsep}{4pt}\renewcommand{\arraystretch}{1.1}
\caption{\textbf{Cross-policy and cross-simulator evaluation of \eco{}.} Scores are percentages unless a unit is shown. $\pm$ is the standard deviation over passes; $\Delta$ rows combine the base and ECO deviations in quadrature. \textit{Top:} HD-Score on the 185-scene HUGSIM test set by domain and overall, with route completion, no-collision, TTC, and comfort. \textit{Bottom:} Evaluation in AlpaSim, an independent simulator with no comfort score; lin./nonlin.\ MPC: AlpaSim's linear/nonlinear model-predictive controller. \eco{} consistently improves the aggregate scores of all base policies in both simulators.}    \label{tab:headline}
    \begin{tabularx}{\textwidth}{@{\hspace{\tabcolsep}} l *{9}{Y} @{\hspace{\tabcolsep}}}
        \toprule
        & \multicolumn{9}{c}{\cellcolor{hugsimhead}\textbf{HUGSIM}~\cite{zhou2024HUGSIM} Test Set (185 scenes), mean over 3 evaluation seeds} \\
        \cmidrule(lr){2-10}
        & \multicolumn{5}{c}{HD-Score$\uparrow$} & & & & \\
        \cmidrule(lr){2-6}
        Policy & Waymo & nuScenes & KITTI-360 & PandaSet & \textbf{Overall} & $R_c\uparrow$ & NC$\uparrow$ & TTC$\uparrow$ & COM$\uparrow$ \\
        \midrule
        VaVAM~\cite{vavam2025} & \bcmp{\nRawPanelWaymo}{\nEcoHPanelWaymo} & \bcmp{\nRawPanelNusc}{\nEcoHPanelNusc} & \bcmp{\nRawPanelKitti}{\nEcoHPanelKitti} & \bcmp{\nRawPanelPanda}{\nEcoHPanelPanda} & \bcmp{\nRawPanelHD}{\nEcoHPanelHD}\sdpm{\nRawPanelHDsd} & \bcmp{\nRawPanelRC}{\nEcoHPanelRC} & \bcmp{\nRawPanelNC}{\nEcoHPanelNC} & \bcmp{\nRawPanelTTC}{\nEcoHPanelTTC} & \bcmp{\nRawPanelCOM}{\nEcoHPanelCOM} \\
        \rowcolor{ecorow} VaVAM + \eco{} & \bcmp{\nEcoHPanelWaymo}{\nRawPanelWaymo} & \bcmp{\nEcoHPanelNusc}{\nRawPanelNusc} & \bcmp{\nEcoHPanelKitti}{\nRawPanelKitti} & \bcmp{\nEcoHPanelPanda}{\nRawPanelPanda} & \bcmp{\nEcoHPanelHD}{\nRawPanelHD}\sdpm{\nEcoHPanelHDsd} & \bcmp{\nEcoHPanelRC}{\nRawPanelRC} & \bcmp{\nEcoHPanelNC}{\nRawPanelNC} & \bcmp{\nEcoHPanelTTC}{\nRawPanelTTC} & \bcmp{\nEcoHPanelCOM}{\nRawPanelCOM} \\
        \midrule
        $\Delta$ (\eco{} $-$ base) & +14.2 & +16.5 & +18.2 & +7.2 & \dEcohPanelVsRawPanel{}$\pm$\sEcohPanelVsRawPanel{} & +4.3 & +2 & +8 & +83 \\
        \midrule
        LTF~\cite{Chitta2023PAMI} & \bcmp{\nLtfRawPanelWaymo}{\nLtfEcoHPanelWaymo} & \bcmp{\nLtfRawPanelNusc}{\nLtfEcoHPanelNusc} & \bcmp{\nLtfRawPanelKitti}{\nLtfEcoHPanelKitti} & \bcmp{\nLtfRawPanelPanda}{\nLtfEcoHPanelPanda} & \bcmp{\nLtfRawPanelHD}{\nLtfEcoHPanelHD}\sdpm{\nLtfRawPanelHDsd} & \bcmp{\nLtfRawPanelRC}{\nLtfEcoHPanelRC} & \bcmp{\nLtfRawPanelNC}{\nLtfEcoHPanelNC} & \bcmp{\nLtfRawPanelTTC}{\nLtfEcoHPanelTTC} & \bcmp{\nLtfRawPanelCOM}{\nLtfEcoHPanelCOM} \\
        \rowcolor{ecorow} LTF + \eco{} & \bcmp{\nLtfEcoHPanelWaymo}{\nLtfRawPanelWaymo} & \bcmp{\nLtfEcoHPanelNusc}{\nLtfRawPanelNusc} & \bcmp{\nLtfEcoHPanelKitti}{\nLtfRawPanelKitti} & \bcmp{\nLtfEcoHPanelPanda}{\nLtfRawPanelPanda} & \bcmp{\nLtfEcoHPanelHD}{\nLtfRawPanelHD}\sdpm{\nLtfEcoHPanelHDsd} & \bcmp{\nLtfEcoHPanelRC}{\nLtfRawPanelRC} & \bcmp{\nLtfEcoHPanelNC}{\nLtfRawPanelNC} & \bcmp{\nLtfEcoHPanelTTC}{\nLtfRawPanelTTC} & \bcmp{\nLtfEcoHPanelCOM}{\nLtfRawPanelCOM} \\
        \midrule
        $\Delta$ (\eco{} $-$ base) & $-$5.0 & +14.1 & +4.5 & $-$1.0 & \dLtfEcohPanelVsLtfRawPanel{}$\pm$\sLtfEcohPanelVsLtfRawPanel{} & +1.6 & $-$6 & $-$3 & 0 \\
        \midrule
        UniAD~\cite{hu2023uniad} & \bcmp{\nUniadRawPanelWaymo}{\nUniadEcoHPanelWaymo} & \bcmp{\nUniadRawPanelNusc}{\nUniadEcoHPanelNusc} & \bcmp{\nUniadRawPanelKitti}{\nUniadEcoHPanelKitti} & \bcmp{\nUniadRawPanelPanda}{\nUniadEcoHPanelPanda} & \bcmp{\nUniadRawPanelHD}{\nUniadEcoHPanelHD}\sdpm{\nUniadRawPanelHDsd} & \bcmp{\nUniadRawPanelRC}{\nUniadEcoHPanelRC} & \bcmp{\nUniadRawPanelNC}{\nUniadEcoHPanelNC} & \bcmp{\nUniadRawPanelTTC}{\nUniadEcoHPanelTTC} & \bcmp{\nUniadRawPanelCOM}{\nUniadEcoHPanelCOM} \\
        \rowcolor{ecorow} UniAD + \eco{} & \bcmp{\nUniadEcoHPanelWaymo}{\nUniadRawPanelWaymo} & \bcmp{\nUniadEcoHPanelNusc}{\nUniadRawPanelNusc} & \bcmp{\nUniadEcoHPanelKitti}{\nUniadRawPanelKitti} & \bcmp{\nUniadEcoHPanelPanda}{\nUniadRawPanelPanda} & \bcmp{\nUniadEcoHPanelHD}{\nUniadRawPanelHD}$\pm$0.8 & \bcmp{\nUniadEcoHPanelRC}{\nUniadRawPanelRC} & \bcmp{\nUniadEcoHPanelNC}{\nUniadRawPanelNC} & \bcmp{\nUniadEcoHPanelTTC}{\nUniadRawPanelTTC} & \bcmp{\nUniadEcoHPanelCOM}{\nUniadRawPanelCOM} \\
        \midrule
        $\Delta$ (\eco{} $-$ base) & +1.0 & +9.5 & +5.2 & +5.4 & \dUniadEcohPanelVsUniadRawPanel{}$\pm$\sUniadEcohPanelVsUniadRawPanel{} & +3.1 & +3 & +6 & +28 \\
        \midrule
        VAD~\cite{jiang2023vad} & \bcmp{\nVadRawPanelWaymo}{\nVadEcoHPanelWaymo} & \bcmp{\nVadRawPanelNusc}{\nVadEcoHPanelNusc} & \bcmp{\nVadRawPanelKitti}{\nVadEcoHPanelKitti} & \bcmp{\nVadRawPanelPanda}{\nVadEcoHPanelPanda} & \bcmp{\nVadRawPanelHD}{\nVadEcoHPanelHD}\sdpm{\nVadRawPanelHDsd} & \bcmp{\nVadRawPanelRC}{\nVadEcoHPanelRC} & \bcmp{\nVadRawPanelNC}{\nVadEcoHPanelNC} & \bcmp{\nVadRawPanelTTC}{\nVadEcoHPanelTTC} & \bcmp{\nVadRawPanelCOM}{\nVadEcoHPanelCOM} \\
        \rowcolor{ecorow} VAD + \eco{} & \bcmp{\nVadEcoHPanelWaymo}{\nVadRawPanelWaymo} & \bcmp{\nVadEcoHPanelNusc}{\nVadRawPanelNusc} & \bcmp{\nVadEcoHPanelKitti}{\nVadRawPanelKitti} & \bcmp{\nVadEcoHPanelPanda}{\nVadRawPanelPanda} & \bcmp{\nVadEcoHPanelHD}{\nVadRawPanelHD}\sdpm{\nVadEcoHPanelHDsd} & \bcmp{\nVadEcoHPanelRC}{\nVadRawPanelRC} & \bcmp{\nVadEcoHPanelNC}{\nVadRawPanelNC} & \bcmp{\nVadEcoHPanelTTC}{\nVadRawPanelTTC} & \bcmp{\nVadEcoHPanelCOM}{\nVadRawPanelCOM} \\
        \midrule
        $\Delta$ (\eco{} $-$ base) & +2.8 & +4.8 & +0.8 & +0.5 & \dVadEcohPanelVsVadRawPanel{}$\pm$\sVadEcohPanelVsVadRawPanel{} & +2.8 & $-$3 & $-$4 & 0 \\
        \midrule
        DrivoR (w/ SimScale~\cite{tian2026simscale})~\cite{kirby2026drivor} & \bcmp{\nDrivorFixRawPanelWaymo}{\nDrivorFixEcoHPanelWaymo} & \bcmp{\nDrivorFixRawPanelNusc}{\nDrivorFixEcoHPanelNusc} & \bcmp{\nDrivorFixRawPanelKitti}{\nDrivorFixEcoHPanelKitti} & \bcmp{\nDrivorFixRawPanelPanda}{\nDrivorFixEcoHPanelPanda} & \bcmp{\nDrivorFixRawPanelHD}{\nDrivorFixEcoHPanelHD}\sdpm{\nDrivorFixRawPanelHDsd} & \bcmp{\nDrivorFixRawPanelRC}{\nDrivorFixEcoHPanelRC} & \bcmp{\nDrivorFixRawPanelNC}{\nDrivorFixEcoHPanelNC} & \bcmp{\nDrivorFixRawPanelTTC}{\nDrivorFixEcoHPanelTTC} & \bcmp{\nDrivorFixRawPanelCOM}{\nDrivorFixEcoHPanelCOM} \\
        \rowcolor{ecorow} DrivoR (w/ SimScale) + \eco{} & \bcmp{\nDrivorFixEcoHPanelWaymo}{\nDrivorFixRawPanelWaymo} & \bcmp{\nDrivorFixEcoHPanelNusc}{\nDrivorFixRawPanelNusc} & \bcmp{\nDrivorFixEcoHPanelKitti}{\nDrivorFixRawPanelKitti} & \bcmp{\nDrivorFixEcoHPanelPanda}{\nDrivorFixRawPanelPanda} & \bcmp{\nDrivorFixEcoHPanelHD}{\nDrivorFixRawPanelHD}\sdpm{\nDrivorFixEcoHPanelHDsd} & \bcmp{\nDrivorFixEcoHPanelRC}{\nDrivorFixRawPanelRC} & \bcmp{\nDrivorFixEcoHPanelNC}{\nDrivorFixRawPanelNC} & \bcmp{\nDrivorFixEcoHPanelTTC}{\nDrivorFixRawPanelTTC} & \bcmp{\nDrivorFixEcoHPanelCOM}{\nDrivorFixRawPanelCOM} \\
        \midrule
        $\Delta$ (\eco{} $-$ base) & +4.9 & +0.8 & +3.5 & $-$0.2 & \dDrivorfixEcohPanelVsDrivorfixRawPanel{}$\pm$\sDrivorfixEcohPanelVsDrivorfixRawPanel{} & +2.0 & $-$1 & 0 & $-$1 \\

    \end{tabularx}

    \vspace{2pt}
    \begin{tabularx}{\textwidth}{@{\hspace{\tabcolsep}} l *{7}{Y} @{\hspace{\tabcolsep}}}
        \toprule
        & \multicolumn{7}{c}{\cellcolor{hugsimhead}\textbf{AlpaSim~\cite{alpasim2025} } NuRec validation set (441 scenes)} \\
        \cmidrule(lr){2-8}
        Policy & \textbf{Scene score}$\uparrow$ & route prog.$\uparrow$ & dist./route$\uparrow$ & m/incident$\uparrow$ & in corridor$\uparrow$ & off-road$\downarrow$ & at-fault coll.$\downarrow$ \\
        \midrule
        VaVAM (nonlin.\ MPC) & 11.0 & 42.9 & 65.6\,m & 143\,m & 57.6 & \textbf{28.0} & 18.0 \\
        \rowcolor{ecorow} VaVAM (nonlin.\ MPC) + \eco{} & \textbf{24.5} & \textbf{51.4} & \textbf{82.4\,m} & \textbf{184\,m} & \textbf{67.1} & 29.7 & \textbf{15.2} \\
        \midrule
        $\Delta$ (\eco{} $-$ base) & +13.5 & +8.5 & +16.8\,m & +41\,m & +9.5 & +1.7 & $-$2.8 \\
        \midrule
        VaVAM (lin.\ MPC) & 32.8 & 60.8 & 112.1\,m & 353\,m & 74.8 & \textbf{15.0} & 16.8 \\
        \rowcolor{ecorow} VaVAM (lin.\ MPC) + \eco{} & \textbf{39.2} & \textbf{63.1} & \textbf{116.0\,m} & \textbf{390\,m} & \textbf{79.8} & 16.8 & \textbf{12.9} \\
        \midrule
        $\Delta$ (\eco{} $-$ base) & +6.4 & +2.3 & +3.9\,m & +37\,m & +5.0 & +1.8 & $-$3.9 \\
        \midrule
        DiffusionDrive~\cite{liao2025diffusiondrive} (nonlin.\ MPC) & 28.9 & 40.6 & 59.8\,m & 190\,m & 77.8 & 22.2 & 9.3 \\
        \rowcolor{ecorow} DiffusionDrive (nonlin.\ MPC) + \eco{} & \textbf{35.2} & \textbf{43.5} & \textbf{67.0\,m} & \textbf{296\,m} & \textbf{81.7} & \textbf{17.1} & \textbf{5.5} \\
        \midrule
        $\Delta$ (\eco{} $-$ base) & +6.3 & +2.9 & +7.2\,m & +106\,m & +3.9 & $-$5.1 & $-$3.8 \\
        \bottomrule
    \end{tabularx}
\end{table*}

%% file: tables/tab_comparison.tex
\begin{table}[t]
    \centering\scriptsize\setlength{\tabcolsep}{3pt}\renewcommand{\arraystretch}{1.1}
    \caption{\textbf{\hd{} on both HUGSIM splits.} Train (345 scenes) and test (185 scenes); $^\star$taken from~\cite{kirby2026drivor}.}
    \label{tab:comparison}
    \begin{tabularx}{\columnwidth}{@{\hspace{\tabcolsep}} l *{6}{Y} @{\hspace{\tabcolsep}}}
        \toprule
        & \multicolumn{2}{c}{VaVAM~\cite{vavam2025}} & \multicolumn{2}{c}{UniAD~\cite{hu2023uniad}} & \multicolumn{2}{c}{\shortstack{DrivoR (w/ SimScale\\\cite{tian2026simscale})~\cite{kirby2026drivor}}} \\
        \cmidrule(lr){2-3}\cmidrule(lr){4-5}\cmidrule(lr){6-7}
        & Train & Test & Train & Test & Train & Test \\
        \midrule
        base policy & \nRawTrainHD & \nRawPanelHD & 32.7$^\star$ & \nUniadRawPanelHD & 35.7$^\star$ & \nDrivorFixRawPanelHD \\
        \rowcolor{ecorow} + \eco{} & \textbf{\nEcoHTrainHD} & \textbf{\nEcoHPanelHD} & \textbf{34.9} & \textbf{\nUniadEcoHPanelHD} & \textbf{\nDrivorFixEcoHTrainHD} & \textbf{\nDrivorFixEcoHPanelHD} \\
        \bottomrule
    \end{tabularx}
\end{table}

%% file: figures/fig_results_overview.tex
\begin{figure}[t]
    \centering
    \includegraphics[width=\columnwidth]{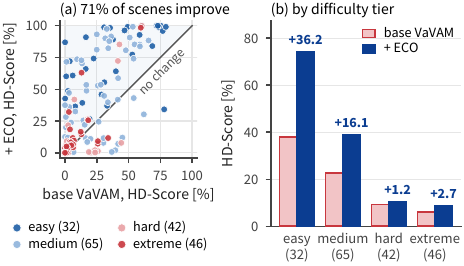}
    \caption{\textbf{Results by HUGSIM difficulty tier.} (a) Every scene of the 185-scene test set as one point: \hd{} without and with \eco{}. Points above the diagonal improve. (b) \eco{} helps most on easier scenes where the base policy is already competent.}
    \label{fig:results}
\end{figure}

%% file: figures/fig_deficit_law.tex
\begin{figure}[t]
    \centering
    \includegraphics[width=\columnwidth]{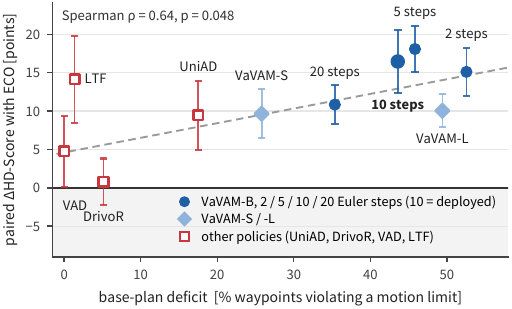}
\caption{\textbf{ECO benefit tends to increase with base-plan deficit.} Paired $\Delta$\hd{} with \eco{} against the fraction of base-plan waypoints that violate a motion limit, for VaVAM-B (\nVamBParams{} parameters) at 2, 5, 10 (deployed), and 20 flow matching steps, VaVAM-S (\nVamSParams{} parameters), VaVAM-L (\nVamLParams{} parameters), and four other policies on the same nuScenes test split in HUGSIM.}
    \label{fig:deficitlaw}
\end{figure}

%% file: tables/tab_endpoint_factorial.tex
\begin{table}[!t]
    \centering\small\setlength{\tabcolsep}{3pt}\renewcommand{\arraystretch}{1.1}
    \caption{{\textbf{Boundary conditions.} VaVAM with the \eco{} objective on the nuScenes split of the HUGSIM test set. Plans are either anchored to the executed history $\mathbf{h}_{-K:0}$ (first predicted waypoint $\bx_0$ free) or pinned at $\bx_0$, with the endpoint $\bx_{N-1}$ either free or pinned.}}
    \label{tab:endpoint}
    \begin{tabular}{l l c c}
        \toprule
        Start boundary & Endpoint $\bx_{N-1}$ & \hd{} & $\Delta$ vs.\ \eco{} \\
        \midrule
        $\bx_0$ pinned & free  & 32.1 & $-$7.5 \\
        history anchored & free & 34.4 & $-$5.2 \\
        $\bx_0$ pinned & pinned  & 36.9 & $-$2.7 \\
        \rowcolor{ecorow} history anchored & pinned  & 39.6 & -- \\
        \midrule
        \multicolumn{2}{c}{base policy} & 23.1 & $-$16.5 \\
        \bottomrule
    \end{tabular}
\end{table}

%% file: figures/fig_operator_gallery.tex
\begin{figure}[!t]
    \centering
    \includegraphics[width=\columnwidth]{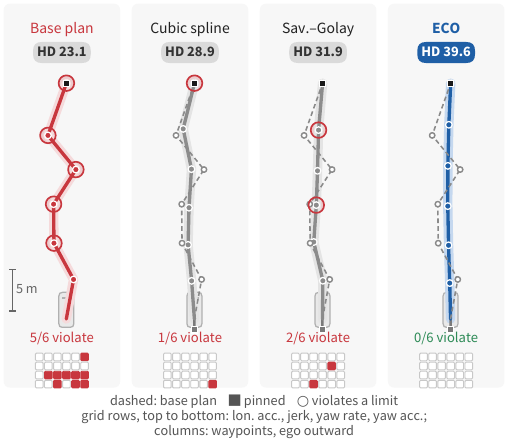}
    \caption{\textbf{Trajectory-shaping operators on a sampled plan.}
All operators are applied with the same boundary conditions as \eco{}: the executed history is anchored, the endpoint is pinned (square), and the first predicted waypoint is free. Rings denote motion-limit violations; HD-Scores are the averages on the nuScenes split (Sec.~\ref{sec:setup}).}
    \label{fig:operator_ladder}
\end{figure}

%% file: sections/06_conclusion.tex
\section{Conclusion}
\label{sec:conclusion}
In this paper, we investigate the OL/CL gap through the executability of a policy's plan, and introduce \eco{}, a lightweight, training-free trajectory optimization layer that keeps the policy's endpoint fixed, anchors the vehicle's executed history, and corrects the trajectory between them. \eco{} yields consistent gains in closed-loop driving across various policies and simulators.
More broadly, our results suggest that closing the OL/CL gap requires considering not only what a driving policy predicts but also whether the waypoint sequence it exposes to the controller is within the motion limits the controller can follow. \eco{} is a simple first step that addresses trajectory-level failures at test time while also complementing methods that can improve the policy's high-level planner.

%% file: sections/07_acknowledgments.tex
\section*{Acknowledgment}
\label{sec:ack}
Claude was used to assist with Matplotlib and LaTeX code generation for Figs.~\ref{fig:teaser}, \ref{fig:pipeline}, \ref{fig:deficitlaw}, and~\ref{fig:operator_ladder}, as well as Python code editing for simulator integration and policy adaptation; all outputs were reviewed and validated by the authors.